\documentclass[10pt,conference]{IEEEtran}
\IEEEoverridecommandlockouts
\usepackage{cite}
\usepackage{amsmath,amssymb,amsfonts}
\usepackage{algorithmic}
\usepackage{graphicx}
\usepackage{textcomp}
\usepackage{xcolor}
\usepackage{CJKutf8}
\usepackage{booktabs}
\usepackage{subfigure} % or subcaption if you're using that
\def\BibTeX{{\rm B\kern-.05em{\sc i\kern-.025em b}\kern-.08em
    T\kern-.1667em\lower.7ex\hbox{E}\kern-.125emX}}
\begin{document}

\title{Comparative Evaluation of Expressive Japanese Character Text-to-Speech with VITS and Style-BERT-VITS2\\
}

% \author{Anonymous Submission}

\author{\IEEEauthorblockN{Zackary Rackauckas}
\IEEEauthorblockA{\textit{Columbia University, RoleGaku} \\
New York, USA \\
zcr2105@columbia.edu}
\and
\IEEEauthorblockN{Julia Hirschberg}
\IEEEauthorblockA{\textit{Columbia University} \\
New York, USA \\
julia@cs.columbia.edu}
}
\maketitle

\begin{abstract}
Synthesizing expressive Japanese character speech poses unique challenges due to pitch-accent sensitivity and stylistic variability. This paper empirically evaluates two open-source text-to-speech models—VITS and Style-BERT-VITS2 JP Extra (SBV2JE)—on in-domain, character-driven Japanese speech. Using three character-specific datasets, we evaluate models across naturalness (mean opinion and comparative mean opinion score), intelligibility (word error rate), and speaker consistency. SBV2JE matches human ground truth in naturalness (MOS 4.37 vs. 4.38), achieves lower WER, and shows slight preference in CMOS. Enhanced by pitch-accent controls and a WavLM-based discriminator, SBV2JE proves effective for applications like language learning and character dialogue generation, despite higher computational demands.
\end{abstract}


\section{Introduction}

Character language represents Japanese speech patterns unique to a fictional character or an archetype of fictional characters \cite{kinsui2015}. Synthesizing speech from character language presents a special set of issues. Distinct character language subsets have varying pitch-accent patterns (see Figure \ref{fig:threecontours}), speaking rates, pause rates, grammar, and vocabulary. Synthesizing speech from in-domain language creates complications in systems producing natural speech with correct phoneme duration, intelligibility, and pitch-accent.

Japanese itself is considered to be the most difficult language for text-to-speech (TTS) models. Pitch-accents are highly context-sensitive, influenced by surrounding words and phrases, and require accurate modeling of prosodic variation, phoneme duration, and character-specific rhythm.

Despite widespread adoption of open-source models like VITS and Style-BERT-VITS2 for character and anime voice synthesis, no formal evaluation exists comparing their performance on expressive Japanese character speech. This work provides the first structured comparative evaluation of these models on in-domain, character-specific datasets. Each model was fine-tuned per character using that character’s voice data, reflecting real-world deployment where consistency and expressive fidelity are critical. While our study does not propose novel architectures, it fills an important empirical gap by evaluating and contrasting model behavior under carefully controlled conditions.

We apply Japanese text-to-speech systems in chatbots for anime-themed language learning applications \cite{rackauckas2024voicedchatbots}. Thus, the naturalness and correctness of TTS models is critical. In this paper, we will focus on two systems: Variational Inference with adversarial learning for end-to-end Text-to-Speech (VITS) \cite{kim2021conditionalvariationalautoencoderadversarial} and Style-enhanced Bidirectional Encoder Representations from Transformers VITS2 (Style-BERT-VITS2) \cite{StyleBertVITS2}. Specifically, we will investigate these systems to answer these questions: (1) Does Style-BERT-VITS2 produce highly accurate audio on in-domain character language? (2) How does VITS compare to Style-BERT-VITS2 in the same production?

We evaluate Style-BERT-VITS2 using a Mean Opinion Score (MOS) compared to a human ground truth benchmark, and compare VITS and Style-BERT-VITS2 models using Comparative Mean Opinion Score (CMOS), intra-speaker consistency, and Word Error Rate (WER). These metrics subjectively measure speech naturalness and objectively measure consistency and intelligibility.

\section{Related Work}

Japanese is a pitch-accent language, and synthesizing correct pitch-accents proves to be a challenge even in modern TTS models \cite{Yasuda_2022, xin2022improvingspeechprosodyaudiobook}. Pitch-accents are highly context-sensitive, influenced by surrounding words and phrases through phenomena such as {\it sandhi} effects \cite{ItoHirose2025}. Recent work has explored contextual approaches to improve pitch-accent prediction. For instance, Hida et al. (2022) demonstrated that combining explicit morphological features with pretrained language models like BERT significantly enhances pitch-accent prediction \cite{hida2022polyphonedisambiguationaccentprediction}. Similarly, pretrained models such as PnG BERT have shown promise in integrating grapheme and phoneme information for better pitch-accent rendering \cite{jia2021pngbertaugmentedbert}. Yasuda and Toda (2022) found that fine-tuning PnG BERT with tone prediction tasks led to notable improvements over baseline Tacotron models \cite{wang2017tacotronendtoendspeechsynthesis} in accent correctness \cite{Yasuda_2022}. However, end-to-end models like Tacotron often struggle with pitch-accent consistency, particularly due to phenomena like accent \textit{sandhi}. Park et al. (2022) highlighted these challenges and proposed a unified accent estimation framework based on multi-task learning to address limitations in end-to-end systems \cite{park22b_interspeech}. These findings emphasize the importance of sophisticated linguistic and contextual modeling.

Accurate phoneme duration prediction is critical for achieving natural speech rhythm and intelligibility, especially in Japanese, where speaking styles and character archetypes introduce variability. Traditional duration models, such as Hidden Markov Models (HMMs), have been widely used in parametric synthesis systems such as OpenJTalk \cite{pyopenjtalk, zen2007hmm}. Modern approaches like FastSpeech2 improve upon this by incorporating variance predictors for explicit control of pitch, energy, and duration \cite{ren2022fastspeech2fasthighquality}. Furthermore, latent duration modeling techniques using vector-quantized variational autoencoder (VAE) to allow for joint optimization of duration and acoustic features, enabling more natural rhythm and timing in synthesized speech \cite{yasuda2020vqvae}. This ability to adaptively predict phoneme duration is particularly important for synthesizing character speech, where exaggerated prosody or unique speech patterns are often required.

Intra-speaker consistency is an important challenge in character language synthesis, as it ensures that a TTS model reliably maintains a speaker's identity across multiple audio samples. This consistency allows speech to remain distinctly recognizable as belonging to the same character. Previous efforts have addressed this through methods like speaker consistency loss, which enforces consistent speaker identity across utterances \cite{makishima2022speakerconsistencylossstepwise}, and adversarial training approaches, which align speaker embeddings using untranscribed data to improve speaker similarity in multi-speaker and zero-shot scenarios \cite{choi2022adversarialspeakerconsistencylearningusing}. Additionally, self-supervised speaker embedding networks have been integrated with TTS models to improve speaker verification and maintain consistent speaker identity in synthesized speech \cite{cho2020learningspeakerembeddingtexttospeech}. More recently, Stable-TTS \cite{han2024stablettsstablespeakeradaptivetexttospeech} introduced a speaker-adaptive framework that uses prior prosody prompting and prior-preservation loss to ensure prosody and timbre consistency, even with limited or noisy target samples. By leveraging high-quality prior samples from pretraining datasets, Stable-TTS addresses issues of variability and overfitting, making it especially valuable for character language synthesis, where a strong and consistent auditory association with a specific persona is essential.

Despite widespread adoption of open-source models like VITS and Style-BERT-VITS2 for character and anime voice synthesis, no formal evaluation exists comparing their performance on expressive Japanese character speech. This work provides the first structured comparative evaluation of these models on in-domain, character-specific datasets. Each model was fine-tuned per character using that character’s voice data, reflecting real-world deployment where consistency and expressive fidelity are critical.

\section{Models}
\subsection{VITS}
VITS is a single-stage text-to-speech framework that employs a variational autoencoder with normalizing flows to model the complex distribution of speech signals. A posterior encoder extracts acoustic features from ground-truth audio, while a prior encoder converts text inputs, typically phonemes, into a latent space refined through normalizing flows. This refined latent representation is then passed on to a HiFi-GAN-style decoder to reconstruct waveforms end-to-end. It uses a flow-based stochastic duration predictor, which models phoneme duration via a learned probability distribution, enabling robust alignment. Additionally, adversarial training is used to enhance the realism of generated waveforms through discriminator networks  \cite{kim2021conditionalvariationalautoencoderadversarial}.

Building on VITS, subsequent advancements have aimed to address limitations in naturalness, efficiency, and expressiveness. VITS2 retained the foundational components of VITS while introducing a transformer block in the normalizing flows to capture long-term dependencies, enhancing the synthesis of coherent and expressive speech. The stochastic duration predictor was refined with adversarial learning, improving efficiency and naturalness \cite{kong2023vits2improvingqualityefficiency}. Furthermore, VITS2 reduced reliance on phoneme conversion by enabling the use of normalized text inputs, marking a significant step toward fully end-to-end TTS systems.  BERT-VITS2 by fishaudio extended VITS2 by integrating a multilingual BERT encoder for text processing, enabling richer contextual understanding and improved text-to-semantic mapping \cite{BertVITS2}. This addition allowed the model to handle diverse linguistic contexts effectively, resulting in more expressive and natural speech synthesis. Unlike BERT-VITS2, Style-BERT-VITS2 incorporates style embeddings, which enable fine-grained emotional and stylistic control \cite{style_bert_vits2_jp_extra_2024}. 

\subsection{Style-BERT-VITS2 JP Extra}

\begin{figure*}[!t]
    \centering
    \subfigure[Hikari F0 Contour \label{fig:sub1}]{
        \includegraphics[width=0.5\linewidth]{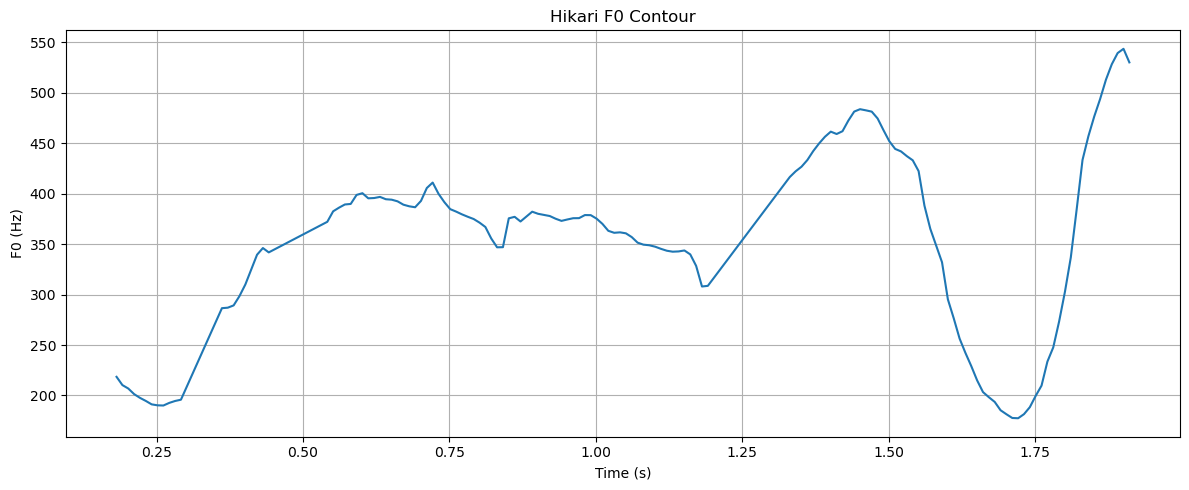}
    }
    \hfill
    \subfigure[Saya F0 Contour \label{fig:sub2}]{
        \includegraphics[width=0.5\linewidth]{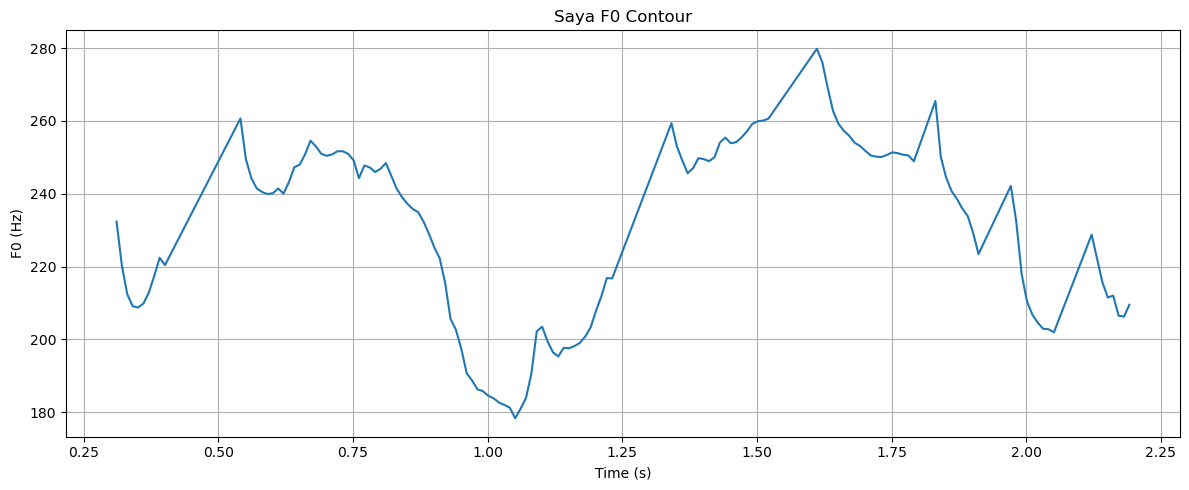}
    }
    \hfill
    \subfigure[Kitsune F0 Contour \label{fig:sub3}]{
        \includegraphics[width=0.5\linewidth]{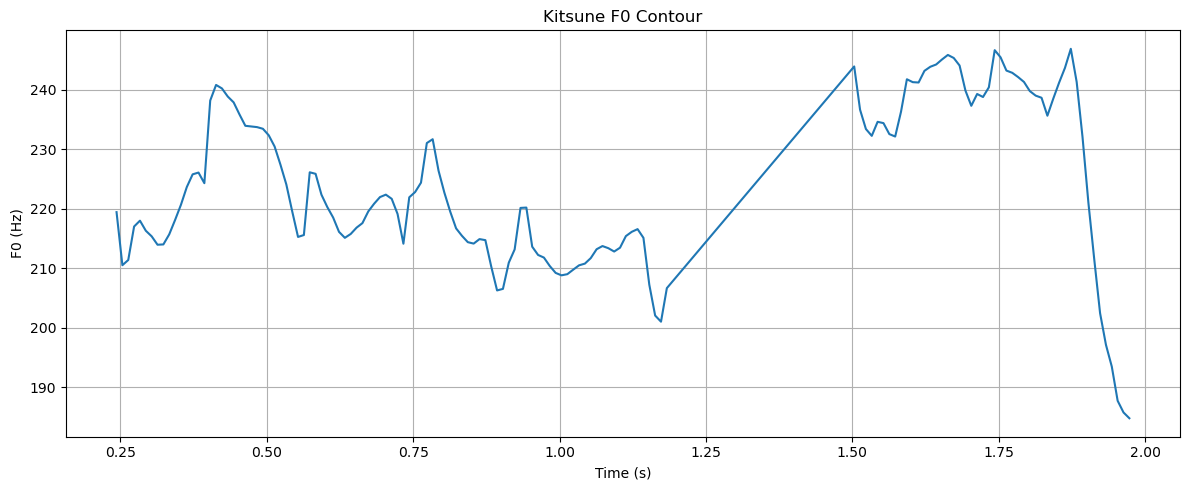}
    }
    \caption{Pitch (F0) contours of self-introduction utterances (“X wa Y desu”) for the three characters. Hikari exhibits a rising contour followed by a sharp fall, Saya shows a mid-level onset with a steep drop and subsequent rise–fall, and Kitsune presents a gradual decline ending in a sharp final drop. These differences highlight the distinct prosodic patterns captured by each character’s voice.}
    \label{fig:threecontours}
\end{figure*}

Style-BERT-VITS2 JP Extra (SBV2JE) \cite{style_bert_vits2_jp_extra_2024}, created by litagin02, enhances the naturalness, expressiveness, and linguistic precision of Japanese TTS synthesis. Building on Style-BERT-VITS2, it addressed challenges in generating native-like Japanese speech, particularly in pitch-accents and phoneme timing. To achieve this, an 800-hour Japanese-only pretraining dataset was utilized alongside architectural modifications, including the addition of a WavLM-based discriminator for enhanced naturalness and the removal of the duration discriminator to stabilize phoneme spacing. Increasing the $gin\_channels$ parameter from 256 to 512 further boosted the model's expressive capacity \cite{style_bert_vits2_jp_extra_2024}.
Fully connected layers were employed to generate style vectors, ensuring precise emotional modulation. Manual pitch-accent adjustment was also introduced, granting users greater control over naturalness in generated speech \cite{style_bert_vits2_2023}. These enhancements make SBV2JE particularly effective for applications such as character-driven speech and emotionally engaging storytelling.

The development of SBV2JE demonstrates a systematic approach to overcoming the challenges of Japanese TTS synthesis. By emphasizing Japanese-specific preprocessing methods, such as g2p (grapheme-to-phoneme) conversions via PyOpenJTalk \cite{pyopenjtalk}, and removing ineffective components like Contrastive Language-Audio Pretraining (CLAP) embeddings \cite{elizalde2022claplearningaudioconcepts}, the model prioritizes linguistic accuracy. This progression establishes SBV2JE as a reference standard for high-quality, expressive, and natural Japanese speech synthesis, catering to applications requiring cultural and emotional adaptability.

Additionally, Figure \ref{fig:threecontours} demonstrates SBV2JE's ability to capture pitch nuances in character language. Figure \ref{fig:sub1} shows the F0 contour of a natural introduction by Hikari, Figure \ref{fig:sub2} of Saya, and Figure \ref{fig:sub3} of Kitsune. Hikari, Kitsune, and Saya are unique characters introduced in Section \ref{sec:method}.
See that Hikari's pitch starts low, rising throughout the sentence, and peaking before showing a large dip, ending at nearly 550 Hz. However, Saya's pitch starts at a middling rate (roughly 230 Hz), exhibits a large drop in the middle of the sentence, then a large rise and a subsequent fall to roughly 200 Hz. Finally, Kitsune's pitch also starts at a middling rate, slowly declining until 1.2 seconds, then peaking and drastically falling off at the end. These patterns visually represent the different pitches capture by different character language sets and different voices in generated audio.

\section{Method}
\label{sec:method}
As outlined in \cite{rackauckas2024voicedchatbots}, audio data was collected from two professional Japanese voice actors using scripts generated by GPT-4 and proofread by a native Japanese speaker to ensure linguistic accuracy and naturalness. Recordings were conducted under conditions to produce high-quality audio without background noise, distortions, or artifacts. The character Hikari was voiced by Voice Actor 1, while Kitsune and Saya were performed by Voice Actor 2. Distinct intonation patterns and phoneme durations were employed to differentiate the characters Kitsune and Saya, despite sharing the same voice actor. The datasets consisted of approximately 10 minutes and 17 seconds of audio for Hikari, 15 minutes and three seconds for Kitsune, and 15 minutes and four seconds for Saya.

Three VITS models were fine-tuned using the vits-fast-finetuning repository \cite{VITSFastFineTuning}. Each was initialized from the pretrained vits-uma-genshin-honkai model, which supports multilingual synthesis for Chinese, Japanese, and English. Fine-tuning was performed individually for Hikari, Kitsune, and Saya, using their respective audio datasets. For preprocessing: Audio files were first denoised using Demucs to isolate vocal components and remove background noise. Stereo audio was converted to mono, and all audio files were resampled to a uniform sampling rate using the torchaudio library, with the target rate specified in the configuration. Long audios were sliced into smaller segments by silence detection with vilero-sad. Transcriptions were generated using the Whisper-large ASR model, producing highly accurate multilingual text alignments. These transcripts were normalized, cleaned, and formatted into the required structure for the VITS training pipeline, where metadata consisted of the audio file path, speaker ID, language ID, and phoneme sequence. Transcripts exceeding 150 characters were excluded to maintain alignment quality. Each dataset was assigned a unique speaker ID.

This per-character fine-tuning approach was essential for evaluating model performance on expressive, character-specific language. Since each voice exhibits unique prosody, pitch-accent patterns, and phoneme timing, a single model trained across characters would not capture character-specific expressiveness or speaker consistency reliably.

The fine-tuning process was configured with 200 epochs, a batch size of 16, and an initial learning rate of 0.0002, decaying at a rate of 0.999875 per step. The Adam optimizer was used with $\beta_1=0.8$, $\beta_2=0.99$, and $\epsilon=1\times10^{-9}$, alongside mixed precision (fp16) for faster training. The model architecture consisted of six Transformer layers with 192 hidden channels, a dropout rate of 0.1, and an upsampling network with kernel sizes $[16, 16, 4, 4]$ and 512 initial channels. This ensured robust learning and enabled high-quality, expressive speech synthesis tailored to the vocal and linguistic styles of Hikari, Kitsune, and Saya. Models were trained on a 40 GB A100 GPU.

In addition, three SBV2JE models were fine-tuned using the official repository. Similar transcription and alignment processes were applied, with audio data sliced and then transcribed using Whisper-large. Two separate preprocessing pipelines were employed to generate BERT-based features and a neutral style vector for conditioning the SBV2JE models. The first pipeline processed text and audio data to generate phoneme-aligned BERT embeddings. Training and validation datasets were formatted with metadata including the audio path, speaker ID, language, text, phoneme sequence, tone, and word-to-phoneme alignment. Phoneme sequences and tone contours were converted into numerical representations and normalized, with optional blank tokens inserted to improve alignment with the acoustic model. BERT embeddings were extracted for each transcription and saved as .bert.pt files. These embeddings were checked to ensure that they matched the phoneme sequence length, with files regenerated if mismatches occurred.

The second pipeline extracted style vectors from audio files using the pretrained Pyannote wespeaker-voxceleb-resnet34-LM model \cite{Wang2023, Bredin23}. Style vectors were saved as .npy files, and their integrity was validated to ensure that no Not a Number (NaN) values were present. Files with NaN values were excluded from the training data to maintain quality. The final training dataset consisted of audio files with valid style vectors, ensuring robust conditioning inputs for the SBV2JE models. These pipelines enabled the synthesis of expressive and stylistically accurate speech outputs for Hikari, Kitsune, and Saya. Because the training datasets only consisted of 10-15 minutes of audio, all audio was used for training, with no audio used for evaluation.

Hikari's model was trained for 200 epochs with a learning rate of 0.0001, while Kitsune's and Saya's were each trained for 400 epochs with a learning rate of 0.00005. The training used a batch size of 4, with the learning rate decaying at a rate of 0.99996 per step. The model architecture consisted of six Transformer layers with 192 hidden channels, a dropout rate of 0.1, and an upsampling network with kernel sizes [16,16,8,2,2] and 512 initial channels. Mel-spectrogram features were extracted with 128 mel channels, a filter length of 2048, a hop length of 512, and a sampling rate of 44,100 Hz. Speaker conditioning used 512 global information (GIN) channels, and the model integrated a WavLM-based discriminator. Models were trained on an A4000 GPU.

\section{Experiments}
\subsection{Mean Opinion Score}
We evaluated the closeness of the SBV2JE models to the ground truth. 11 crowd-sourced native Japanese speakers listened to 60 audios each and provided ratings on a one through five Likert scale. Participants were instructed to rate based on naturalness of speech, appropriate phoneme duration, and correctness of intonation and pitch-accent. Audios were chosen through random stratified sampling to ensure an equal number of audios for each model and ground truth and a sufficient variation of vocabulary and sentence structures.
\renewcommand{\arraystretch}{1.3} % Optional: add row spacing

\begin{table}[th]
  \caption{MOS Results for TTS and Ground Truth Audio}
  \label{tab:mos_results}
  \centering
  {\normalsize  % change to \normalsize or \large for more boost
  \begin{tabular}{ l  r@{ $\pm$ }l }
    \toprule
    \textbf{Speaker-Type} & \multicolumn{2}{c}{\textbf{MOS $\pm$ SD}} \\
    \midrule
    Hikari (SBV2JE) & 4.47 & 0.68 \\
    Hikari (GT) & 4.76 & 0.48 \\
    Kitsune (SBV2JE) & 4.40 & 0.78 \\
    Kitsune (GT) & 4.14 & 0.89 \\
    Saya (SBV2JE) & 4.24 & 0.76 \\
    Saya (GT) & 4.24 & 0.75 \\
    \midrule
    \textbf{Overall TTS} & 4.37 & 0.74 \\
    \textbf{Overall GT} & 4.38 & 0.77 \\
    \bottomrule
  \end{tabular}
  }
\end{table}

All models performed comparably to the ground truth in terms of naturalness. Hikari TTS achieved a score of $4.47 \pm 0.68$, slightly underperforming relative to its ground truth counterpart ($4.76 \pm 0.48$). This indicated that raters generally perceived the ground truth audio as more natural than the TTS-generated audio for Hikari. Conversely, Kitsune TTS slightly outperformed its ground truth, with scores of $4.40 \pm 0.78$ and $4.14 \pm 0.89$, respectively. This suggested that raters found the TTS-generated audio to be more natural than the ground truth for Kitsune, indicating that certain aspects of the synthesized audio surpassed the quality of the ground truth recordings. Finally, Saya TTS and Saya ground truth were nearly identical, receiving scores of $4.24 \pm 0.76$ and $4.24 \pm 0.75$, respectively, suggesting parity in perceived naturalness. Kitsune ground truth exhibited the highest variability ($\pm 0.89$), despite being recorded by a professional voice actor. This highlighted subjective differences in rater perceptions. Moreover, the overall score for TTS ($4.37 \pm 0.74$) was nearly identical to the overall ground truth score ($4.38 \pm 0.77$). A two-tailed, paired t-test on aggregate rating data yielded an insignificant result with $p \approx 0.91$, indicating that native speakers did not perceive a statistically significant difference in naturalness between the SBV2JE audio and the ground truth recordings.

\subsection{Comparative Mean Opinion Score}
To evaluate the difference in human preference between VITS and SBV2JE, we used Comparative Mean Opinion Score with 5 native Japanese speakers as raters. Raters were presented two audios generated using the same text, one from VITS and one from SBV2JE, and were asked to rate their preference on a scale of -3 (greatly prefer the second model) to 3 (greatly prefer the first model). The position of models presented was randomized in addition to the speakers. Ten audios were used from each speaker, resulting in a total of 150 ratings.
\renewcommand{\arraystretch}{1.3} % Optional: increases row height

\begin{table}[th]
  \caption{CMOS Results for TTS Models}
  \label{tab:cmos_results}
  \centering
  {\normalsize % Change to \normalsize or \large for slightly bigger
  \begin{tabular}{ l  r }
    \toprule
    \textbf{Speaker} & \textbf{CMOS} \\
    \midrule
    Hikari & -0.12 \\
    Saya   & -0.06 \\
    Kitsune & -0.03 \\
    \midrule
    \textbf{Overall} & -0.06 \\
    \bottomrule
  \end{tabular}
  }
\end{table}

The comparative mean opinion score analysis indicated that SBV2JE was slightly preferred over VITS, with an overall CMOS of -0.06. When evaluating individual characters, Hikari showed a CMOS of -0.12, Saya had a CMOS of -0.06, and Kitsune presented a CMOS of -0.03, all favoring SBV2JE. These results suggested that, while both models performed comparably, SBV2JE had a slight advantage in perceived quality across the evaluated characters.

\subsection{Intra-Speaker Consistency}
We calculated intra-speaker similarity by leveraging speaker embeddings generated by the pretrained Resemblyzer model \cite{Resemblyzer}. Audio files were grouped by speaker and model, and each file was processed to extract embeddings representing the speaker's identity. For each speaker-model pair, we computed cosine similarity between all pairs of embeddings, which quantified how consistently the model maintained the same speaker's identity across multiple audio samples. The average similarity score for each speaker-model pair was calculated, and these scores were further grouped by speaker and model to evaluate trends. This approach provided a robust measure of how well each TTS model preserved speaker consistency.
\renewcommand{\arraystretch}{1.3} % Increases vertical spacing between rows

\begin{table}[th]
  \caption{Intra-Speaker Similarity Scores for TTS Models}
  \label{tab:intra_speaker_similarity}
  \centering
  {\normalsize % You can switch to \normalsize or \large if you want it a bit bigger
  \begin{tabular}{ l  r }
    \toprule
    \textbf{Speaker-Model} & \textbf{Similarity Score} \\
    \midrule
    Hikari (VITS) & 0.96 \\
    Saya (VITS) & 0.96 \\
    Hikari (SBV2JE) & 0.96 \\
    Kitsune (VITS) & 0.97 \\
    Kitsune (SBV2JE) & 0.97 \\
    Saya (SBV2JE) & 0.96 \\
    \midrule
    \textbf{Grouped by Speaker} & \textbf{Average Similarity} \\
    Hikari & 0.96 \\
    Saya & 0.96 \\
    Kitsune & 0.97 \\
    \midrule
    \textbf{Grouped by Model} & \textbf{Average Similarity} \\
    VITS & 0.96 \\
    SBV2JE & 0.96 \\
    \bottomrule
  \end{tabular}
  }
\end{table}

Both the VITS and SBV2JE models achieved high intra-speaker consistency across all speakers, with similarity scores exceeding 95\% in all cases. Among the speakers, Kitsune exhibited the highest average similarity score (97.05\%), indicating that the models handled this speaker's identity particularly well. In contrast, Saya showed slightly lower consistency (95.78\%), suggesting that maintaining this speaker's identity was more challenging for the models. When grouped by model, VITS slightly outperformed SBV2JE (96.43\% vs. 96.24\%), although the difference was minimal. These results highlighted the robustness of both models in preserving speaker identity, with only minor variations across speakers and models. This strong performance is encouraging for real-world applications where consistent speaker representation is essential, such as in character-driven TTS systems for language learning. Further analysis could explore how differences in speaker characteristics can contribute to variability.

\subsection{Word Error Rate}
We calculated Word Error Rate to assess intelligibility by comparing generated speech against corresponding ground truth text. First, we transcribed the audio files using the Faster-Whisper ASR model \cite{fasterwhisper}, initialized with a seed prompt for better context. To ensure linguistic accuracy, we tokenized both the transcriptions and ground truth text using Sudachi, a Japanese morphological analyzer \cite{TAKAOKA18.8884}. For each transcription, we identified the most relevant ground truth line using Levenshtein similarity matching, ensuring a fair comparison even if minor differences existed between the audio content and ground truth. Finally, WER was calculated by measuring the proportion of word-level errors (insertions, deletions, and substitutions) between the tokenized transcription and the matched ground truth. This methodology provided a robust evaluation of how intelligible and accurate the TTS outputs were when processed through a real-world ASR system.
\renewcommand{\arraystretch}{1.3} % Optional: adds vertical padding between rows

\begin{table}[th]
  \caption{Word Error Rate Comparison}
  \label{tab:wer_comparison}
  \centering
  {\normalsize % You can bump to \normalsize or \large if needed
  \begin{tabular}{ l  r }
    \toprule
    \textbf{Speaker-Model} & \textbf{WER (Mean $\pm$ SD)} \\
    \midrule
    Hikari (VITS) & 0.03 $\pm$ 0.03 \\
    Hikari (SBV2JE) & 0.03 $\pm$ 0.02 \\
    Saya (VITS) & 0.08 $\pm$ 0.05 \\
    Saya (SBV2JE) & 0.08 $\pm$ 0.07 \\
    Kitsune (VITS) & 0.03 $\pm$ 0.04 \\
    Kitsune (SBV2JE) & 0.02 $\pm$ 0.02 \\
    \midrule
    \textbf{Grouped by Model} & \textbf{Average WER} \\
    VITS & 0.05 $\pm$ 0.04 \\
    SBV2JE & 0.04 $\pm$ 0.04 \\
    \bottomrule
  \end{tabular}
  }
\end{table}

Both the VITS and SBV2JE models achieved high intelligibility across all speakers, with consistently low WER values. Kitsune showed the best performance, with SBV2JE achieving the lowest WER (0.02), while Saya had the highest WER (0.08) for both models. When grouped by model, SBV2JE slightly outperformed VITS in overall WER (0.04 vs. 0.05) with comparable variability. These results highlighted both models' reliability, with SBV2JE offering slightly greater consistency, making them well-suited for character-driven TTS applications in language learning.

\section{Discussion}
Our experiments confirm SBV2JE as a highly natural and intelligible Japanese TTS model, achieving slight but meaningful improvements over VITS in WER and CMOS. These gains stem from key architectural modifications, including the removal of the duration discriminator, which stabilizes phoneme spacing for smoother speech, and the integration of a WavLM-based discriminator, enhancing prosodic nuance and expressiveness. Tailored for Japanese-specific challenges, SBV2JE incorporates pitch-accent adjustments and g2p conversions while streamlining the architecture by removing ineffective components like CLAP embeddings, emphasizing the value of domain-specific optimization in advancing TTS quality.

While these improvements make SBV2JE well-suited for character-driven conversational agents and language learning applications, the model can be applied to  storytelling, anime voiceover, personalized voice assistants, and other applications where naturalness and expressiveness are critical. 

In addition to their use in anime-focused language learning tools, expressive TTS models like VITS and Style-BERT-VITS2 are well-suited for live storytelling, where emotional nuance plays a key role in delivering a compelling narrative. Although our work focuses on Japanese, both models support English and Chinese as well, opening the door to cross-lingual applications such as voice translation and code-switching across these languages. These models can also be deployed as backend APIs for mobile or edge devices, broadening their practical utility across platforms. Finally, speech synthesis connects users through the Internet of Things (IoT) \cite{zhao2023emotion} and personalized voice assistance.

However, there are limitations to consider. For example, SBV2JE presents  trade-offs between model complexity and computational efficiency. While the model’s enhancements yield measurable gains, the increased computational cost associated with its architecture, such as higher gin channel values and WavLM-based discriminators, could pose challenges for deployment on resource-constrained devices. 

While untested in this study, the model's control over stylistic elements could open up opportunities for more adaptive and engaging user experiences in interactive platforms. These stylistic elements open future work in fine-grained emotion control, real-time conversational adaptation, cross-lingual expressive synthesis, personalized voice tutoring for language learners, and low-resource style transfer for underrepresented character voices.

Our study is constrained by the small per-character datasets, the limited number of characters and raters, and the SBV2JE repository’s use of all available audio for training, and while these factors restrict generalization and efficiency analysis, our evaluations focus on perceptual judgments of generated speech, leaving broader validation to future work.

These results highlight SBV2JE’s practical advantages for real-world expressive Japanese TTS systems, particularly where character identity and pitch correctness are essential.

\section{Conclusion}
SBV2JE represents a significnt advancement in Japanese TTS character language synthesis, balancing naturalness, intelligibility, and expressiveness. By addressing linguistic nuances and leveraging innovative architectural modifications, it sets a reference standard for character-driven speech synthesis and language learning. Future research should aim to refine its generalization, explore new use cases, and optimize its deployment to maximize its impact across diverse applications.

\section*{Acknowledgments}
This research was initiated independently of the primary author’s academic responsibilities at Columbia University and was conducted without the use of university resources.

\begin{CJK}{UTF8}{min}
\bibliographystyle{IEEEtran}
\bibliography{sample_base}
\end{CJK}
\end{document}